# Lip-Reading Driven Deep Learning Approach for Speech Enhancement

Ahsan Adeel, Mandar Gogate, Amir Hussain, William M. Whitmer

*Abstract*—This paper proposes a novel lip-reading driven deep learning framework for speech enhancement. The proposed approach leverages the complementary strengths of both deep learning and analytical acoustic modelling (filtering based approach) as compared to recently published, comparatively simpler benchmark approaches that rely only on deep learning. The proposed audio-visual (AV) speech enhancement framework operates at two levels. In the first level, a novel deep learning based lip-reading regression model is employed. In the second level, lip-reading approximated clean-audio features are exploited, using an enhanced, visually-derived Wiener filter (EVWF), for the clean audio power spectrum estimation. Specifically, a stacked long-short-term memory (LSTM) based lip-reading regression model is designed for clean audio features estimation using only temporal visual features (i.e. lip reading) considering different number of prior visual frames. For clean speech spectrum estimation, a new filterbank-domain EVWF is formulated, which exploits estimated speech features. The proposed EVWF is compared with conventional Spectral Subtraction (SS) and Log-Minimum Mean-Square Error (LMMSE) methods using both ideal AV mapping and LSTM driven AV mapping. The potential of the proposed speech enhancement framework is evaluated under four different dynamic real-world commercially-motivated scenarios (cafe, street junction, public transport (BUS), pedestrian area) at different SNR levels (ranging from low to high SNRs) using benchmark Grid and ChiME3 corpora. For objective testing, perceptual evaluation of speech quality (PESQ) is used to evaluate the quality of restored speech. For subjective testing, the standard mean-opinion-score (MOS) method is used with inferential statistics. Comparative simulation results demonstrate significant lip-reading and speech enhancement improvement in terms of both speech quality and speech intelligibility. Ongoing work is aimed at enhancing the accuracy and generalization capability of the deep learning driven lip-reading model, and contextual integration of AV cues, for context-aware autonomous AV speech enhancement.

## I. Introduction

Speech enhancement aims to enhance the perceived overall speech quality and intelligibility, when the noise degrades them significantly. Due to the extensive speech enhancement requirement in a wide-range of real-world applications, such as mobile communication, speech recognition, hearing aids etc.,

Corresponding author: Amir Hussain (http://www.cs.stir.ac.uk/ ahu/.html).
Ahsan Adeel, Mandar Gogate, Amir Hussain are with the Department of Computing Science and Mathematics, Faculty of Natural Sciences, University of Stirling, UK. (E-mail: {aad, mgo, ahu}@cs.stir.ac.uk)
William M. Whitmer is with MRC/CSO Inst. Of Hearing Research Scottish Section, Glasgow, G31 2ER, UK. (E-mail: william.whitmer@nottingham.ac.uk)

several speech enhancement methods have been proposed over the past few decades, ranging from state-of-the-art statistical, analytical, and classical optimization approaches to advanced deep learning based methods. The classic speech enhancement methods are mainly based on audio only processing such as SS [1], audio-only Wiener filtering [2], minimum mean-square error (MMSE) [3][4], linear minimum mean square error (LMMSE) etc. Recently, researchers have also proposed deep learning based advanced speech recognition [5] and enhancement [6] methods. However, most of these methods are based on single channel (audio only) processing, which often perform poorly in adverse conditions, where overwhelming noise is present [7].

Human speech processing is inherently multimodal, where visual cues help to better understand the speech. The multimodal nature of speech is well established in literature, and it is understood how speech is produced by the vibration of the vocal folds and configuration of the articulatory organs. The correlation between the visible properties of the articulatory organs (e.g., lips, teeth, tongue) and speech reception has been previously shown in numerous behavioural studies [8][9][10][11]. Therefore, the clear visibility of some articulatory organs could be effectively utilized to approximate a clean speech signal out of a noisy audio background. The biggest advantage of using visual cues to extract clean audio features is their inherent noise immunity: the visual speech representation always remains unaffected by acoustic noise [12].

In the recent literature, extensive research has been carried out to develop multimodal speech processing methods, which establishes the importance of multimodal information in speech processing [13]. Researchers have proposed novel visual feature extraction methods [14][15][16][17][18], fusion approaches (early integration [19], late integration [20], hybrid integration [21]), multi-modal datasets [22][23], and fusion techniques [21][24][25]. The multimodal audiovisual speech processing methods have shown significant performance improvement [26]. Moreover, with the advent of advanced and more sophisticated digital signal processing approaches (in terms of both hardware and software), researchers have shown some ground breaking performance improvements. For example, the authors in [27] proposed a novel deep learning based lip-reading system with 93% accuracy. The authors in [28] proposed a multimodal hybrid deep neural network (DNN) architecture based on deep convolutional neural network (CNN) and Bidirectional Long Short-Term Memory (BiLSTM) network for speech enhancement. Similarly, the authors in [29] proposed an audiovisual speech recogni-tion system, where deep learning algorithms, such as CNN, deep denoising autoencoders, and multistream hidden Markov model (MSHMM) models have been used for visual feature extraction, audio feature extraction, and audiovisual features integration respectively.

Recently, the authors in [30] proposed an audio-visual speech enhancement approach using multimodal deep CNNs. Specifically, the authors developed an audio-visual deep CNN (AVDCNN) speech enhancement model that integrates audio and visual cues into a unified network model. The proposed AVDCNN approach is structured as an audio-visual encoder-decoder where both audio and visual cues are processed using two separate CNNs, and later fused into a joint network to generate enhanced speech. However, the proposed AVDCNN approach relies only on deep CNN models. For testing, the authors used self-prepared dataset that contained video recordings of 320 utterances of Mandarin sentences spoken by a native speaker (only one speaker). In addition, the used noises include usual car engine, baby crying, pure music, music with lyrics, siren, one background talker (1T), two background talkers (2T), and three background talkers (3T). In summary, for testing only one Speaker (with 40 clean utterances mixed with 10 noise types at 5 dB, 0 dB, and -5 dB SIRs) was considered.

In contrast to [30], our proposed novel approach leverages the complementary strengths of both deep learning and analytical acoustic modelling (filtering based approach). For testing, we used five different speakers including two white males, two white females, and one black male speaker (that ensured speaker independence criteria) with real-world dynamic noises in extreme noisy scenarios. Specifically, we propose a novel deep learning based lip-reading regression model and EVWF for speech enhancement. The proposed speech enhancement framework first reads target speakers lips and estimates clean audio features using stacked LSTM model. Afterwards, the estimated clean audio features are fed into the EVWF for speech enhancement. The proposed lip-reading driven EVWF approach effectively exploits temporal correlation in lip-movements by considering different number of prior visual frames. The presented EVWF model is tested with the challenging benchmark AV ChiME3 corpus in real-world commercially-motivated scenarios (such as cafe, street junction, public transport (BUS), pedestrian areas) for SNRs ranging from -12 to 12dB. Comparative simulation results demonstrate that the proposed EVWF approach outperforms benchmark audio-only approaches at very low SNRs (-12dB, -6dB, -3dB, and 0dB), and the improvement is statistically significant at the 95% confidence level.

The four major contributions presented in this paper are:

1) Proposed a novel deep learning based lip-reading regression model for speech enhancement applications. Specifically, a stacked LSTM based data-driven model is proposed to approximate the clean audio features using only temporal visual features (i.e. lip reading). In the literature, extensive research has been carried out to model lip reading as a classification problem for speech recognition. In contrast, not much work has been conducted to model lip reading as a regression problem for speech enhancement [27][28][29].
2) A critical analysis of the proposed LSTM based lip-reading regression model and its comparison with the conventional MLP based regression model [31], where LSTM model has shown better capability to learn the correlation between lip movements and speech as compared to the conventional MLP models, particularly, when different number of prior visual frames are considered.
3) Addressed limitations of state-of-the-art VWF by presenting a novel EVWF. The proposed EVWF has employed an inverse filter-bank (FB) transformation (i.e. a pseudoinverse of the approximated audio features) for audio power spectrum estimation as compared to the cubic spline interpolation method (that fails to estimate the missing power spectral values when it interpolates the low dimensional vector to high dimensional audio vector and ultimately leads to a poor audio power spectrum estimation). In addition, the proposed EVWF has eliminated the need for voice activity detection (VAD) and noise estimation.
4) Evaluated the potential of our proposed speech enhancement framework, exploiting both ideal AV mapping (ideal visual to audio feature mapping) and designed stacked LSTM based lip-reading model. The benchmark AV Grid and ChiME3 corpora are employed, with 4 different real-world noise types (cafe, street junction, public transport (BUS), pedestrian area), to objectively and subjectively evaluate the performance of the EVWF.

The rest of the paper is organized as follows: Section II presents the proposed AV speech enhancement framework, designed EVWF, and lip-reading algorithms. Section III presents the employed AV dataset and audiovisual feature extraction methodology. Section IV presents comparative experimental results, including the evaluation of deep learning-driven AV mapping and EVWF based speech enhancement. Finally, Section V concludes this work and proposes some future research directions.

## II. Speech Enhancement Framework

The proposed two-level deep learning based lip reading driven speech enhancement framework is depicted in Fig. 1. In the first level, a novel lip-reading regression model is effectively utilized to estimate clean audio features using only temporal visual features. In the second level, estimated low dimensional clean audio features are transformed into high dimensional clean audio power spectrum using inverse FB transformation to calculate Wiener filter. Finally, the Wiener filter is applied to the magnitude spectrum of the noisy input audio signal, followed by the inverse fast Fourier transform (IFFT), overlap, and combining processes to produce enhanced magnitude spectrum. The state-of-the-art VWF and designed EVWF are depicted in Fig.2. and Fig. 3 respectively. It is to be noted that the designed EVWF has addressed the limitations of state-of-the-art VWF [12] by employing an inverse FB transformation (i.e. a pseudoinverse of the approximated audio features) for audio power spectrum estimation as compared to the cubic spline interpolation method. In addition, the designed EVWF has eliminated the need for VAD and noise estimation.

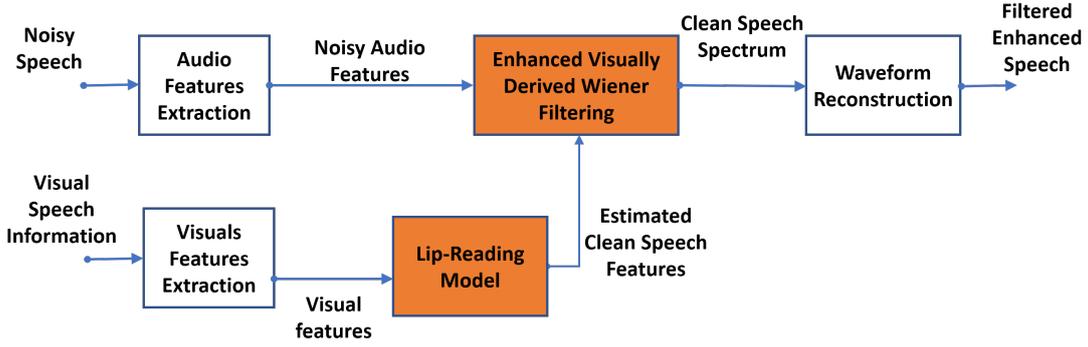

Fig. 1: Proposed lip-reading driven deep learning approach for speech enhancement. The system first estimates clean audio features using visually derived speech model (i.e. lip reading). Afterwards, the estimated clean audio features are fed into the proposed enhanced visually derived Wiener filter for the estimation of clean speech spectrum.

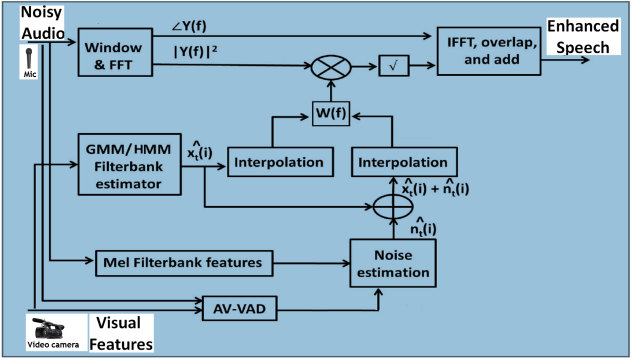

Fig. 2: State-of-the-art visually-derived Wiener filtering [12]. The authors in [12], presented a hidden Markov model-Gaussian mixture model (HMM/GMM) based two-level state-of-the-art VWF for speech enhancement. However, the use of HMM/GMM models for the estimation of clean audio features ($\hat{X_t}(i)$) from visual features and cubic spline interpolation for the approximation of high dimensional clean audio power spectrum from the estimated low dimensional audio features are not optimal choices. The HMM/GMM model suffers from poor generalization and cubic spline interpolation method fails to estimate the missing power spectral values that leads to a poor audio power spectrum estimation.

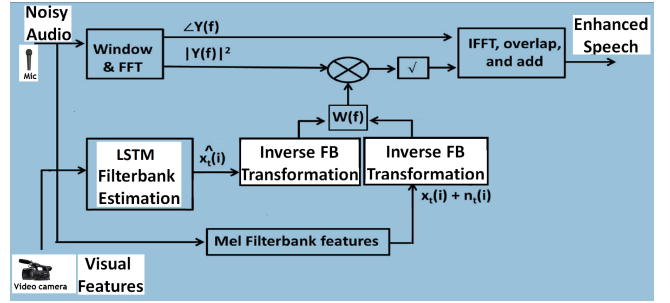

Fig. 3: Proposed enhanced visually-derived Wiener filtering. Note the use of LSTM based FB estimation and an inverse FB transformation for audio power spectrum estimation. The proposed approach addressed both the power spectrum estimation and generalization issues of the state-of-the-art VWF. In addition, it replaced the need of voice activity detector and noise estimator.

### A. Enhanced Visually Derived Wiener Filter

In signal processing, Wiener filter is a state-of-the-art filter that helps to produce an estimate of a clean audio signal by linear time-invariant (LTI) filtering of an observed noisy audio signal. The frequency domain Wiener Filter is defined as:

$$W(\gamma) = \frac{\psi\hat{a}(\gamma)}{\psi_a(\gamma)} \quad (1)$$

where $\psi_a(\gamma)$ is the noisy audio power spectrum (i.e. clean power spectrum + noisy power spectrum) and $\psi\hat{a}(\gamma)$ is the clean audio power spectrum. The calculation of the noisy audio power spectrum is fairly easy because of the available noisy audio vector. However, the calculation of the clean audio power spectrum is challenging which restricts the use of Wiener filter widely. Hence, for successful Wiener filtering, it is necessary to acquire the clean audio power spectrum. In this paper, the clean audio power spectrum is calculated using deep learning based lip reading driven speech model.

The FB domain Wiener filter ($\hat{W}_t^{FB}(k)$) is given as [12]:

$$\hat{W}_t^{FB}(k) = \frac{\hat{x}_t(k)}{\hat{x}_t(k) + \hat{n}_t(k)} \quad (2)$$

where $\hat{x}_t(k)$ is the FB domain lip-reading driven approximated clean audio feature and $\hat{n}_t(k)$ is the FB domain noise signal. The subscripts $k$ and $t$ represents the $k^{th}$ channel and $t^{th}$ audio frame.

The lip-reading driven approximated clean audio feature vector $\hat{x}_t(k)$ is a low dimensional vector. For high dimensional Wiener filter calculation, it is necessary to transform the estimated low dimensional FB domain audio coefficients into a high dimensional power spectral domain. It is to be noted that the approximated clean audio and noise features $(\hat{x}_t(k) + \hat{n}_t(k))$ are replaced with the noisy audio (a combination of real clean speech $(x_t(k))$ and noise $(n_t(k))$). The low dimension to high dimension transformation can be written as:

$$\hat{W}^{FB}_{t_{[N_l,M]}}(k) = \frac{\hat{x}_t(k)_{[N_l,M]}}{x_t(k)_{[N_l,M]} + n_t(k)_{[N_l,M]}} \quad (3)$$

$$\hat{W}^{FB}_{t_{[N_h,M]}}(k) = \frac{\hat{x}_t(k)_{[N_h,M]}}{x_t(k)_{[N_h,M]} + n_t(k)_{[N_h,M]}} \quad (4)$$

Where $N_l$ and $N_h$ are the low and high dimensional audio features respectively, and M is the number of audio frames. The transformation from (3) to (4) is very crucial to the performance of filtering. Therefore, the authors in [12] proposed the use of cubic spline interpolation method to determine the missing spectral values. However, the use of cubic spline interpolation for the approximation of high dimensional clean audio power spectrum from a low dimensional audio FB features is not an optimal approach. The interpolation based method fails to estimate the missing power spectral values. In contrast, this article proposes the use of inverse FB transformation which used the least square pseudo-inverse based method to approximate the optimal high dimensional power spectrum.

The inverse FB domain transformation is calculated as follows:

$$\hat{x}_{t(k)_{[N_h,M]}} = \hat{x}_{t(k)_{[N_l,M]}} * \alpha_x \quad (5)$$

$$n_{t(k)_{[N_h,M]}} = n_{t(k)_{[N_l,M]}} * \alpha_n \quad (6)$$

$$\alpha_x = \alpha_n = (\phi_m(k)^T \phi_m(k))^{-1} \phi_m(k)^T \quad (7)$$

$\phi_m(k) =$
$$\begin{cases} 0 & k < f_{b_{mf-1}} \\ \frac{k - f_{b_{mf-1}}}{f_{b_{mf}} - f_{b_{mf-1}}} & f_{b_{mf-1}} \leq k \leq f_{b_{mf}} \\ \frac{f_{b_{mf-1}} - k}{f_{b_{mf+1}} + f_{b_{mf}}} & f_{b_{mf}} \leq k \leq f_{b_{mf+1}} \\ 0 & k > f_{b_{mf+1}} \end{cases}$$

where $f_{b_{mf}}$ are the boundary points of the filters and corresponds to the $k^{th}$ coefficient of the $k$-points DFT. The boundary points $f_{b_{mf}}$ are calculated using:

$$f_{b_{mf}} = (\frac{K}{F_{samp}} \cdot f_{cm_f}) \quad (8)$$

where $f_{cm_f}$ is the mel scale frequency

After substitutions, the obtained $k$-bin Wiener filter (4) is then applied to the magnitude spectrum of the noisy audio signal (i.e. $|Y_t(k)|$) to estimate the enhanced magnitude audio spectrum ($|\hat{X}_{t(k)_{[N_H,M]}}|$). The enhanced magnitude audio spectrum is given as:

$$|\hat{X}_{t(k)}| = |Y_{t(k)}||\hat{W}_{t_{[N_h,M]}}| \quad (9)$$

The acquired time-domain enhanced speech signal (9) is then followed by the IFFT, overlap, and combining processes.

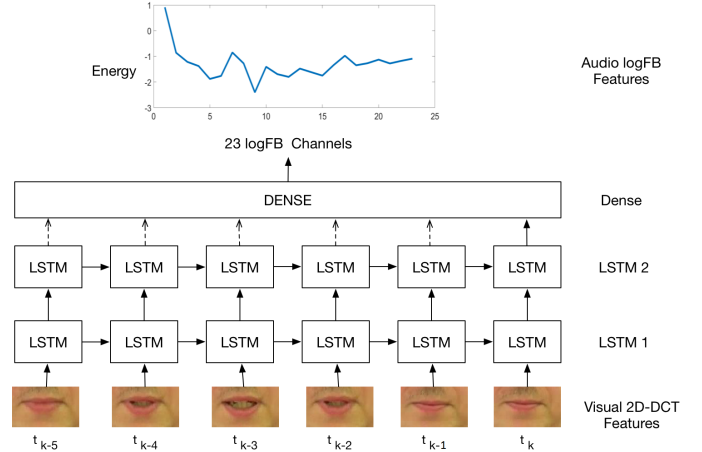

Fig. 4: Stacked long-short-term memory based lip reading regression model: Example of 5 prior visual frames (taking into account the current visual frame $t_k$ as well as the temporal information of previous visual frames $t_{k-1}, t_{k-2}, t_{k-3}, t_{k-4}, t_{k-5}$)

### B. Lip Reading Regression Model

*1) Long-Short-Term Memory:* This section describes the LSTM network architecture summarised in Fig. 4. LSTM was originally proposed in [32] by Sepp Hochreiter and Jrgen Schmidhuber. The LSTM network consists of input layer, two LSTM layers, and output dense layer. Visual features of time instance $t_k, t_{k-1}, ..., t_{k-5}$ ($k$ is the current time instance and 5 is the number of prior visual frames) were fed into the stacked LSTM layers. The lower LSTM layer has 250 cells, which encoded the input and passed its hidden state to the second LSTM layer, which has 300 cells. The output of the second LSTM layer was then fed into the fully connected (dense) layer which has total 23 neurons with linear activation function. In this architecture, the input at layer $k$ is the value of hidden state $h_t$ computed by layer $k-1$. The stacked LSTM architecture was trained with the objective to minimise the mean squared error (MSE) between the predicted and the actual audio features. The MSE (10) between the estimated audio logFB features and clean audio features was minimised using stochastic gradient decent algorithm and RMSProp optimiser. RMSprop is an adaptive learning rate optimizer which divides the learning rate by moving average of the magnitudes of recent gradients to make learning more efficient. Moreover, to reduce the overfitting, dropout (of 0.25) was applied after every LSTM layer. The MSE cost function $C(a_{estimated}, a_{clean})$ can be written as:

$$C(a_{\text{estimated}}, a_{\text{clean}}) = \sum_{i=1}^{n} 0.5(a_{\text{estimated}}(i) - a_{\text{clean}}(i))^2 \quad (10)$$

where $a_{estimated}$ and $a_{clean}$ are the estimated and clean audio features respectively.

*2) Multilayer perceptron:* MLPs are made up of highly interconnected processing elements called as neurons, processes the information by their state response and learn from examples. A neuron in an MLP is connected to several inputs with different associated weights. The output of a neuron is the summation of all connected inputs, followed by a non-linear processing unit, called as a transfer function. The main objective of MLP is to transform the inputs into meaningful outputs, learn the input-output relationship, and offer viable solutions to unseen problems (a generalization capability). Therefore, the capacity to learn from examples is one of the most desirable features of neural network models. The goal of training is to learn desired system behaviour and adjust the network parameters (interconnections weights) to map (learn) the input-output relationship and minimize the cost function. The processed audiovisual corpus was fed into an MLP model in order to learn the relationship between lip-movements and speech signal. The MSE cost function is given in (10). For training different number of hidden neurons ranging from 10 to 150 were used, with 1 and 2 hidden layers. More details are comprehensively presented in Section IV.

TABLE I: Summary of sentences from the Grid Corpus

| Speaker ID | Grid ID | Total | Full | | Aligned | |
|---|---|---|---|---|---|---|
| | | | Removed | Used | Removed | Used |
| Speaker 1 | S1 | 1000 | 11 | 989 | 11 | 989 |
| Speaker 2 | S15 | 1000 | 164 | 836 | 164 | 836 |
| Speaker 3 | S26 | 1000 | 16 | 984 | 71 | 929 |
| Speaker 4 | S6 | 1000 | 9 | 991 | 9 | 991 |
| Speaker 5 | S7 | 1000 | 11 | 989 | 11 | 989 |

*C. Input & Output preprocessing*

Both LSTM and MLP networks ingest visual discrete cosine transform (DCT) features of time instance $t_n, t_{n-1}, ..., t_{n-k}$, where $n$ is the current time instance and $k$ is the number of prior visual frames as shown in Fig. 4. The Input layer of the networks are organised such that at $k^{th}$ time step LSTM/MLP receives temporal input. The output of the dense layer is logFB audio feature.

III. DATASET AND AUDIOVISUAL FEATURE EXTRACTION

*A. Dataset*

In this paper, a well-established Grid [22] and ChiME3 [33] corpora are used. The clean Grid videos are mixed with ChiME3 noises (cafe, street junction, public transport (BUS), pedestrian area) for SNRs ranging from -12 to 12dB to develop a new AV ChiME3 corpus. The preprocessing includes sentence alignment and incorporation of prior visual frames. The sentence alignment is performed to remove the silence time from the video and prevent model from learning redundant or insignificant information. The preprocessing enforced the model to learn the correlation between the spoken word and corresponding visual representation, rather than over learning the silence. Secondly, prior multiple visual frames are used to incorporate temporal information to improve mapping between visual and audio features. The Grid corpus comprised of 34 speakers each speaker reciting 1000 sentences. Out of 34 speakers, a subset of 5 speakers is selected (two white females, two white males, and one black male) with total 900 command sentences each. The subset fairly ensures the speaker independence criteria. A summary of the acquired visual dataset is presented in Table I, where the full and aligned sentences, total number of sentences, used sentences, and removed sentences are clearly defined. The audio and visual features extraction procedure is depicted in Fig. 5.

*B. Audio feature extraction*

The audio features are extracted using widely used log-FB vectors. For log-FB vectors calculation, the input audio signal is sampled at 50kHz and segmented into $N$ 16ms frames with 800 samples per frame and 62.5% increment rate. Afterwards, a hamming window and Fourier transformation is applied to produce 2048-bin power spectrum. Finally, a 23-dimensional log-FB is applied, followed by the logarithmic compression to produce 23-D log-FB signal.

*C. Visual feature extraction*

The visual features are extracted from the Grid Corpus videos recorded at 25 fps using a 2D-DCT based standard and widely used visual feature extraction method. Firstly, the video files are processed to extract a sequence of individual frames. Secondly, a Viola-Jones lip detector [34] is used to identify lip-region by defining the Region-of-Interest (ROI) in terms of bounding box. Object detection is performed using Haar feature-based cascade classifiers. The method is based on machine learning where cascade function is trained with positive and negative images. Finally, the object tracker [35] is used to track the lip regions across the sequence of frames. The visual extraction procedure produced a set of corner points for each frame, where lip regions are then extracted by cropping the raw image. In addition, to ensure good lip tracking, each sentence is manually validated by inspecting few frames from each sentence. The aim of manual validation is to delete those sentences in which lip regions are not correctly identified [31]. Lip tracking optimization lies outside the scope of the present work. The development of a full autonomous system and its testing on challenging datasets is in progress.

In the final stage of visual features extraction, the 2D-DCT of lip region is calculated to produce vectors of pixel intensities. Afterwards, to produce the final frame vector, first 50 components are vectorized in a zigzag order and then the DCT features are interpolated to match the equivalent audio sequence. It is to be noted that videos are recorded at 25 fps and corresponding audio files are produced at 75 vectors per second (VPS). Therefore, the produced visual vectors are upsampled to match the 75 VPS rate. This is performed by copying each visual vector three times and mapped to three consecutive audio frames (e.g. V1, V1, V1 → A1, A2, A3).

IV. EXPERIMENTAL RESULTS

*A. Methodology*

For audiovisual mapping and estimation of clean audio features, both MLP and LSTM based data-driven approaches are used. Firstly, in Section IV. B, an MLP based learning model is trained and validated for initial data analysis. The datasets analysis with MLP revealed that the selected dataset possesses good enough speakers variability to demonstrate the potential of our proposed approach and that it is optimal to use aligned sentences. In section IV. C, both MLP and LSTM models are tested and compared using multiple visual frames for enhanced AV mapping. It is to be noted that different MLP architectures are used as compared to LSTM. It is mainly due to MLP's uncertainty and dependence on the number of neurons, whereas LSTM's dependency is not at the same degree. Therefore, LSTM's optimized structure was predicted without many structural trials. A subset of dataset is used to train the neural network (80% training dataset) and rest of the data (20%) is used to test and validate the performance of the trained neural network in face of new context (10% validation, 10% testing). Table II summarizes the Train, Test, and Validation sentences. For generalization testing, the proposed framework is trained on SNRs ranging from -10 to 12dB, and tested on -12dB SNR. The neural network performances are evaluated using MSE with the goal to achieve the least possible MSE. The estimated lip-reading driven audio features are exploited by the designed novel EVWF for speech enhancement. The speech enhancement of our proposed EVWF is compared with the state-of-the-art SS and Log MMSE based audio-only speech enhancement approaches. For noisy speech, clean utterances were mixed with noisy backgrounds (randomly chosen from ChiME3 noises) to produce noisy utterances of SNRs ranging from -12dB to 12dB.

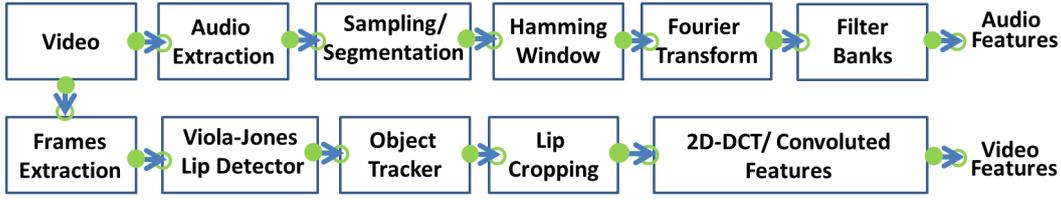

Fig. 5: Audiovisual dataset generation procedure

TABLE II: Summary of train, test, and validation sentences from Grid corpus

| Speakers | Train | Validation | Test | Total |
|---|---|---|---|---|
| 1 | 692 | 99 | 198 | 989 |
| 2 | 585 | 84 | 167 | 836 |
| 3 | 650 | 93 | 186 | 929 |
| 4 | 693 | 99 | 199 | 991 |
| 5 | 692 | 99 | 198 | 989 |
| All | 3312 | 474 | 948 | 4734 |

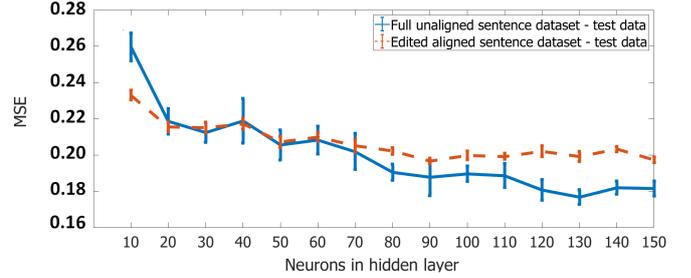

Fig. 6: MSE median test data results of using full (solid blue line) and aligned (red dashed line) sentences for different MLP hidden layer sizes. MLP is used for initial data analysis to evaluate unaligned and aligned sentences. The analysis revealed that learning model with full sentences learns silence more because of its over representation.

### B. Initial Data Analysis with MLP

*1) Sentence Length and 1 to 1 Visual to Audio Mapping Evaluation using MLP:* In this subsection, the use of aligned sentences is justified. For sentence length evaluation, full and aligned sentences of all five speakers with 900 sentences each are used. The audio and visual features are extracted and shuffled randomly. The sentence length evaluation results are shown in Fig. 6. It can be seen that the learning model performs better with full sentences as compared to the aligned sentences. However, a closer inspection revealed that the learning model with full sentences over learns silence because of the silence representation. In addition, with the full sentence case, the learning model works as a VAD, where it effectively distinguishes between the silent and speech frames. However, in this article, we aim to learn the relationship between speech and visuals for speech mapping instead of modelling VAD. Therefore, aligned dataset is of more relevance even if it shows a higher MSE.

*2) Evaluation of Speaker Independence Criteria:* In this subsection, the aim is to justify the use of 5 speakers subset for the evaluation of proposed approach. Fig. 7 shows both speaker dependent and independent audiovisual mapping results with MLP. It can be seen that different speakers have achieved different MSE results. The variations in MSE is because of different speakers articulation that leads to good or bad audiovisual mapping; hence, satisfies speaker independence criteria.

### C. Multiple Visual Frames to Audio Mapping using MLP and LSTM

In multiple visual frames to audio mapping, multiple prior frames are used (ranging from 1 visual frame to 27 prior visual frames). The simulation results are shown in Figs. 8 and 9 and Table III. The training is performed with aligned six different datasets (i.e 1, 2, 4, 8, 14, and 18 prior visual frames). The datasets varied in total number of visual frames (including current and prior visual frames). The multiple prior visual frames correspondingly increased the total number of inputs. However, the output dimensions remained same. It can be seen that by moving from 1 visual frame to 18 visual frames, a significant performance improvement could be achieved. The LSTM model with 1 visual frame achieved the MSE of 0.092, whereas with 18 visual frames, the model achieved the least MSE of 0.058. In contrast, the MLP based lip reading model could only achieve the MSE of 0.199 and 0.209 with 1 and 18 visual frames respectively. It is to be noted that MLP remained deficient in achieving low MSE, because MLP architecture lacks the capability of exploiting prior visual frames for better learning. In contrast, LSTM based learning model exploited the temporal information (i.e. prior visual frames) effectively and showed consistent reduction in MSE while going from 1 to 18 visual frames. This is mainly because of its inherent recurrent architectural property and the ability of retaining state over long time spans by using cell gates. The estimated log-FB vectors based on the visual inputs are shown in Fig 10, with both MLP and LSTM models. LSTM's enhanced visual to audio mapping as compared to MLP is evident.

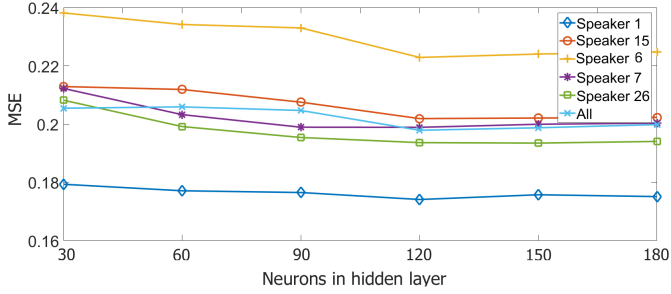

Fig. 7: MSE test data results of using individual and combined speakers (for 1 to 1 audiovisual mapping). MLP is used for initial data analysis to evaluate the fairness of 5 speakers subset and the speaker independence criteria. The simulation results revealed that different speakers have achieved different MSEs due to different speakers articulation (that leads to good or bad audiovisual mapping).

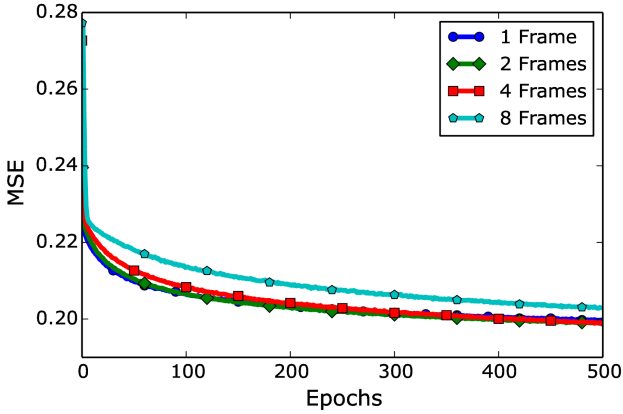

Fig. 8: AV mapping: MLP validation results for different visual frames - All Speakers. The figure presents an overall behaviour of an MLP model when contextual information (i.e. previous frames) is added. It is to be noted that MLP failed to acquire significant performance improvement upon contextual information integration.

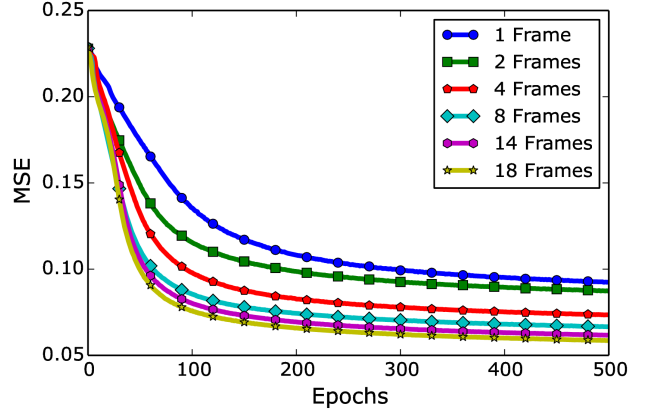

Fig. 9: AV mapping: Stacked LSTM validation results for different visual frames - All Speakers. The figure presents an overall behaviour of an LSTM model when contextual information (i.e. previous frames) is added. It is to be noted that LSTM better exploited the temporal correlation as compared to MLP. However, LSTM saturates at 18 prior visual frames.

TABLE III: MLP vs. LSTM: Training and testing accuracy comparison for different visual frames

| Visual Frames | LSTM | | MLP | |
|---|---|---|---|---|
| | $MSE_{train}$ | $MSE_{test}$ | $MSE_{train}$ | $MSE_{test}$ |
| 1 | 0.092 | 0.104 | 0.199 | 0.204 |
| 2 | 0.087 | 0.097 | 0.199 | 0.202 |
| 4 | 0.073 | 0.085 | 0.198 | 0.200 |
| 8 | 0.066 | 0.082 | 0.202 | 0.204 |
| 14 | 0.061 | 0.080 | 0.210 | 0.218 |
| 18 | 0.058 | 0.078 | 0.217 | 0.209 |

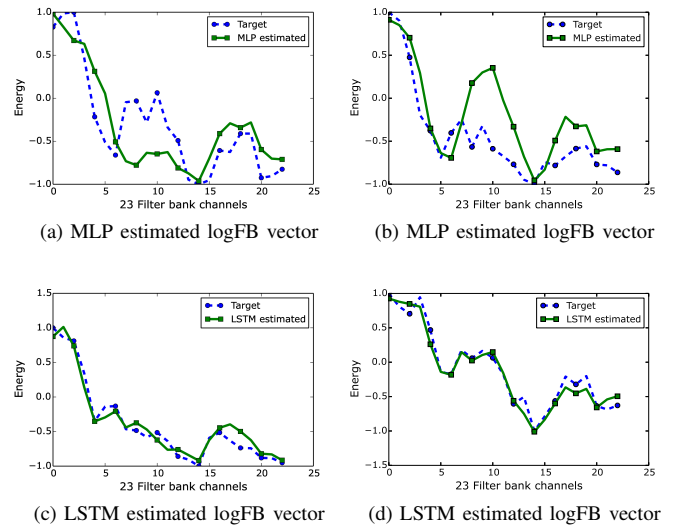

(a) MLP estimated logFB vector  (b) MLP estimated logFB vector
(c) LSTM estimated logFB vector  (d) LSTM estimated logFB vector

Fig. 10: Estimated clean audio features using 14 prior visual frames

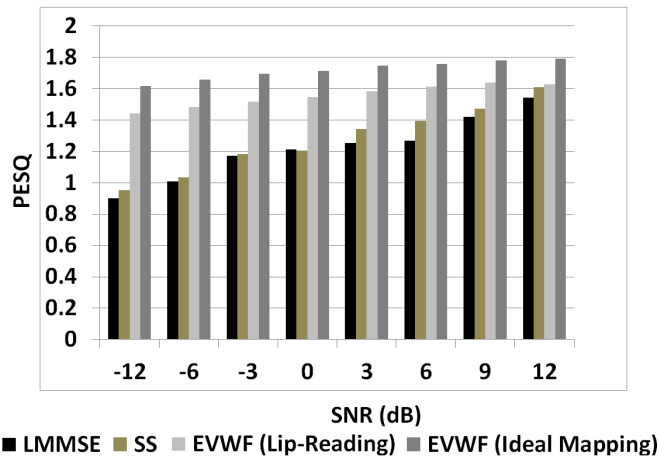

Fig. 11: PESQ with ChiME3 Noises (cafe, street junction, public transport (BUS), pedestrian area). It can be seen that, at low SNR levels, EVWF significantly outperformed both SS and LMMSE based speech enhancement methods. The low PESQ score with ChiME3 corpus, particularly at high SNRs, can be attributed to the nature of the ChiMe3 noise, characterized by spectro-temporal variation, potentially reducing the ability of enhancement algorithms to restore the signal.

### D. Speech Enhancement

*1) Objective Test:* For objective testing, perceptual evaluation of speech quality (PESQ) is used to evaluate the quality of restored speech using AV ChiME3 corpus. The PESQ score is computed as a linear combination of the average disturbance value and the average asymmetrical disturbance values. The PESQ score ranges from -0.5 to 4.5 corresponding to low to high speech quality. The PESQ scores for EVWF, SS, and LMMSE for different SNRs are depicted in Figure 11. It can be seen that, at low SNR, EVWF significantly outperformed both SS [36] and LMMSE [4] based speech enhancement methods. The low PESQ score with ChiME3 corpus, particularly at high SNRs, can be attributed to the nature of the ChiMe3 noise. The latter is characterized by spectro-temporal variation, potentially reducing the ability of enhancement algorithms to restore the signal. Fig. 12 displays the spectrogram of a randomly selected utterance from AV ChiME3 corpus, where the performance of EVWF at very low SNR (-12db) is evident. It is to be noted that for generalization testing, -12dB SNR utterances were not included in the training dataset.

*2) Subjective Listening Tests:* To examine the effectiveness of the proposed EVWF, subjective listening tests were conducted in terms of MOS with self-reported normal-hearing listeners using AV ChiME3 corpus. The listeners were presented with a single stimulus (i.e. enhanced speech only) and were asked to rate the re-constructed speech on a scale of 1 to 5. The five rating choices were: (5) Excellent (when the listener feels unnoticeable difference compared to the target clean speech) (4) Good (perceptible but not annoying) (3) Fair (slightly annoying) (2) Poor (annoying), and (1) Bad (very annoying). The EVWF is compared with two state-of-the-art speech enhancement methods (SS and LMMSE). A total of 10 listeners took part in the evaluation session. The clean speech signal was corrupted with ChiME3 noises (at SNRs of -12dB, -6dB, -3dB, 0dB, 3dB, 6dB, and 12dB). Fig. 13 depicts the performances of three different speech enhancement methods in terms of MOS with ChiME3 noises. In addition, the performance of proposed stack LSTM based lip-reading model is compared with the ideal AV mapping, showing good quality estimation. It can be seen that the proposed AV approach significantly outperformed benchmark Audio-only approaches at low SNRs. At high SNRs, for the case of (spectro-temporally) correlated ChiME3 noises, the AV performed comparably to Audio-only.

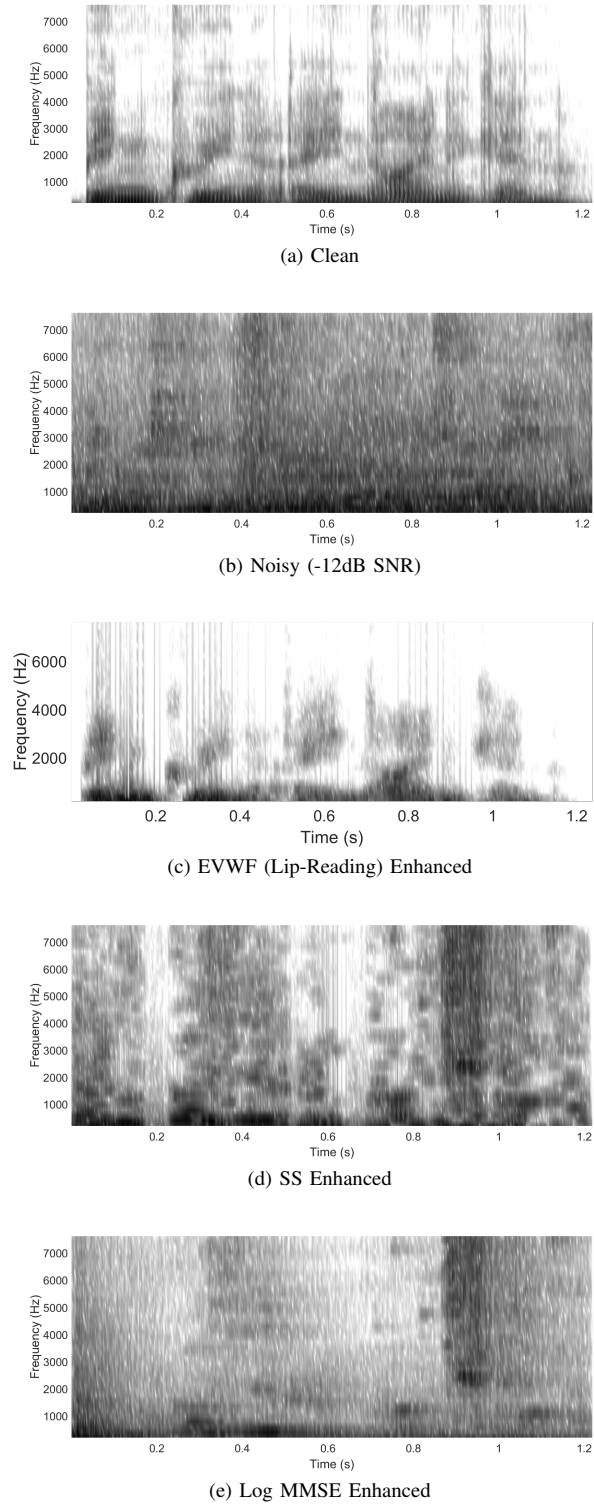

Fig. 12: Spectrogram of a random utterance from AV ChiME3 corpus: (a) Clean (b) Noisy (-12dB SNR) (c) EVWF (Lip-Reading) Enhanced (d) Spectral Subtraction Enhanced (e) Log MMSE Enhanced

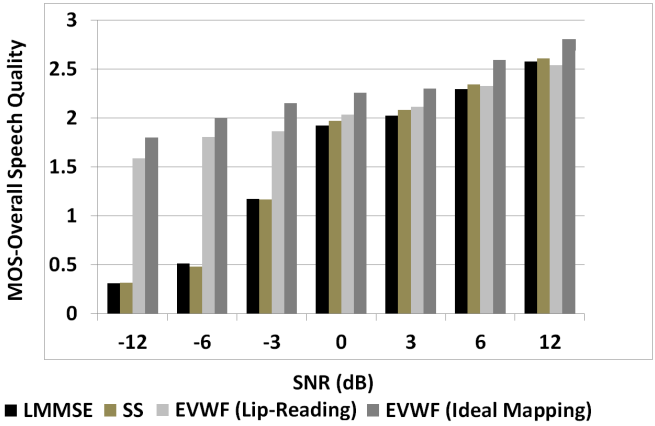

Fig. 13: MOS for overall speech quality with ChiME3 Noises (cafe, street junction, public transport (BUS), pedestrian area). It can be seen that, EVWF significantly outperformed both SS and LMMSE based speech enhancement methods at low SNR.

To further, determine the significance of our results, we compared the performance of our proposed EVWF with the state-of-the-art speech enhancement algorithms using t-test at a significance level of 0.05, following statistical analysis with the t-test approach presented in [37]. For each of the pair-wise comparisons, the null hypothesis $H_0 : \mu1=\mu2$ is defined and tested (whether MOS for two methods is significantly different or not for each SNR). The results of the t-test at the level of 0.05 significance are presented in Tables IV (SS vs. EVWF) and V (Log-MMSE vs. EVWF) for AV ChiME3 dataset. It can be seen that the proposed EVWF significantly outperformed both SS and LMMSE at low SNRs (-12dB, -6dB, and -3dB).

## V. DISCUSSION AND CONCLUSION

In this paper, a novel LSTM based lip-reading regression model is first developed, as part of our proposed audio-visual speech enhancement framework. In addition, a novel filter-bank-domain EVWF is formulated and integrated with a lip reading model, and compared with SS and LMMSE methods. The proposed EVWF employs an inverse filter-bank transformation for audio power spectrum estimation, as compared to the cubic spline interpolation method used by the state-of-the-art VWF model. In addition, the proposed EVWF eliminates the need for VAD and noise estimation. The performance evaluation of the lip-reading model demonstrates

TABLE IV: The results of the t-test at 5% Significance Level (MOS)-ChiME3: Comparison of EVWF with SS. The proposed EVWF under real noisy environments suggests that the proposed AV approach outperforms benchmark SS approach at low SNRs, and the improvement is statistically significant at the 95% confidence level.

| SNRs | p-value | $H_0$: Null hypothesis |
|---|---|---|
| -12dB | 5.23E-05 | (+) |
| -6dB | 6.23E-05 | (+) |
| -3dB | 6.61E-05 | (+) |
| 0dB | 0.2218 | (-) |
| 3dB | 0.3795 | (-) |
| 6dB | 0.9333 | (-) |
| 12dB | 0.6102 | (-) |

TABLE V: The results of the t-test at 5% Significance Level (MOS)-ChiME3: Comparison of EVWF with LMMSE. The proposed EVWF under real noisy environments suggests that the proposed AV approach outperforms benchmark LMMSE approach at low SNRs, and the improvement is statistically significant at the 95% confidence level.

| SNRs | p-value | $H_0$: Null hypothesis |
|---|---|---|
| -12dB | 4.16E-05 | (+) |
| -6dB | 1.26E-04 | (+) |
| -3dB | 4.93E-04 | (+) |
| 0dB | 0.1149 | (-) |
| 3dB | 0.2038 | (-) |
| 6dB | 0.7556 | (-) |
| 12dB | 0.7183 | (-) |

LSTM's enhanced capability to estimate clean audio features as compared to feed forward neural network models, especially when different number of prior visual frames are considered. Comparative performance evaluation of the proposed EVWF under real noisy environments suggests that the proposed AV approach outperforms benchmark Audio-only approaches at low SNRs, and the improvement is statistically significant at the 95% confidence level. Even at high SNRs, the proposed EVWF performs comparably to the conventional speech enhancement approaches. The aforementioned limitation of our proposed AV approach (i.e visual cues become fairly less effective at high SNRs for speech enhancement) leads us to propose future development of a more optimal, context-aware AV system, that can effectively account for different noisy conditions and contextually utilize both visual and noisy audio features. Our ongoing and future work thus aims to develop a context-aware AV integration algorithm to better deal with different noisy environments, and further enhance the accuracy and generalization capability of the current deep learning driven lip-reading model. For generalization testing, this study utilized an adequate (initial) subset of 5 diverse subjects (two white females, two white males, and one black male) with a total of 900 command sentences per speaker. The subset satisfies the speaker independence criteria depicted in Fig. 7, where different speakers can be seen to have achieved different MSEs due to different speakers articulation (that in turn, leads to good or bad AV mapping). The discussion about what constitutes an adequate number of speakers, including testing different datasets, noises etc., requires more work, and will be addressed as a separate topic in a future paper. If we compare our subjective results with recent benchmark works, e.g. Hou et al., 2018 [30], we have conducted more extensive testing in terms of number of speakers and benchmark speech datasets (e.g. Grid and ChiME) at extremely challenging low SNR levels (up to -12dB). The focus of this work was to demonstrate the potential of our proposed novel AV approach. Our contribution is a first of its kind that leverages the complementary strengths of deep learning and analytical acoustic modelling (filtering based) approach as compared to recently published deep learning based approaches. In the future, we intend to investigate the performance of our proposed EVWF under more realistic scenarios, including generalization testing with unfamiliar speakers, use of novel visual features, and deep speech enhancement components. Whilst the preliminary comparative results reported in this paper should be taken with care, they demonstrate the potential and capabilities of our developed deep learning-driven AV approach. What is now needed is further extensive, comparative evaluation against other benchmark speech enhancement approaches, using a range of real, noisy AV corpora. We are currently recording the latter in real conversational settings, and plan to make these available, as new benchmark resources, to the multidisciplinary speech research community.

ACKNOWLEDGMENT

This work was supported by the UK Engineering and Physical Sciences Research Council (EPSRC) Grant No. EP/M026981/1 (CogAVHearinghttp://cogavhearing.cs.stir.ac.uk). In accordance with EPSRC policy, all experimental data used in the project simulations is available at http://hdl.handle.net/11667/81. The authors would like to gratefully acknowledge Dr Andrew Abel from Xi'an Jiaotong-Liverpool University for building and providing the audio-visual dataset. In addition, the authors would like to acknowledge Ricard Marxer and Jon Barker from the University of Sheffield, Roger Watt from the University of Stirling, and Peter Derleth from Sonova, AG, Staefa, Switzerland for their contributions. The authors would also like to thank Kia Dashtipour for conducting MOS test. Lastly, we gratefully acknowledge the support of NVIDIA Corporation for donating the Titan X Pascal GPU for this research.